\documentclass[10pt,twocolumn,letterpaper]{article}

\usepackage{3dv}
\usepackage{times}
\usepackage{epsfig}
\usepackage{graphicx}
\usepackage{amsmath}
\usepackage{amssymb}
\usepackage{textcomp}

\usepackage[pagebackref=true,breaklinks=true,letterpaper=true,colorlinks,bookmarks=false]{hyperref}

\threedvfinalcopy 

\def\threedvPaperID{210} 

\newcommand{\degree}[1]{${#1}^o$}
\def\360{\degree{360}}

\usepackage{xcolor}
\definecolor{yUpColor}{RGB}{127, 127, 255}
\definecolor{yDownColor}{RGB}{208, 207, 4}
\definecolor{zNeg}{RGB}{127, 0, 127}
\definecolor{zPos}{RGB}{127, 255, 127}
\definecolor{xNeg}{RGB}{0, 127, 127}
\definecolor{xPos}{RGB}{255, 127, 127}

\usepackage{caption}
\usepackage{subcaption}
\usepackage{multirow}
\ifthreedvfinal\fi
\begin{document}

\title{\360 Surface Regression with a Hyper-Sphere Loss}

\author{%
    Antonis Karakottas%
\quad \quad%
    Nikolaos Zioulis%
\quad \quad%
    Stamatis Samaras%
\quad \quad%
    Dimitrios Ataloglou\\
\quad \quad%
    Vasileios Gkitsas%
\quad \quad%
    Dimitrios Zarpalas%
\quad \quad%
    Petros Daras\\
\fontsize{11pt}{2pt}\selectfont Centre for Research and Technology Hellas (CERTH) - Information Technologies\\ \fontsize{11pt}{2pt}\selectfont Institute (ITI) - Visual Computing Lab (VCL), Thessaloniki, Greece\\
{\tt\small{\{ankarako, nzioulis, sstamatis, ataloglou, gkitsasv, zarpalas, daras\}@iti.gr}}\\
\small{\url{vcl.iti.gr}}
}%

\maketitle
\thispagestyle{empty}

\begin{abstract}
   Omnidirectional vision is becoming increasingly relevant as more efficient \360 image acquisition is now possible. However, the lack of annotated \360 datasets has hindered the application of deep learning techniques on spherical content. This is further exaggerated on tasks where ground truth acquisition is difficult, such as monocular surface estimation. While recent research approaches on the 2D domain overcome this challenge by relying on generating normals from depth cues using RGB-D sensors, this is very difficult to apply on the spherical domain. In this work, we address the unavailability of sufficient \360 ground truth normal data, by leveraging existing 3D datasets and remodelling them via rendering. We present a dataset of \360 images of indoor spaces with their corresponding ground truth surface normal, and train a deep convolutional neural network (CNN) on the task of monocular \360 surface estimation. We achieve this by minimizing a novel angular loss function defined on the hyper-sphere using simple quaternion algebra. We put an effort to appropriately compare with other state of the art methods trained on planar datasets and finally, present the practical applicability of our trained model on a spherical image re-lighting task using completely unseen data by qualitatively showing the promising generalization ability of our dataset and model. The dataset is available at: \url{vcl3d.github.io/HyperSphereSurfaceRegression}.
\end{abstract}

\section{Introduction}
\begin{figure}
    \centering
    \includegraphics[width = \linewidth]{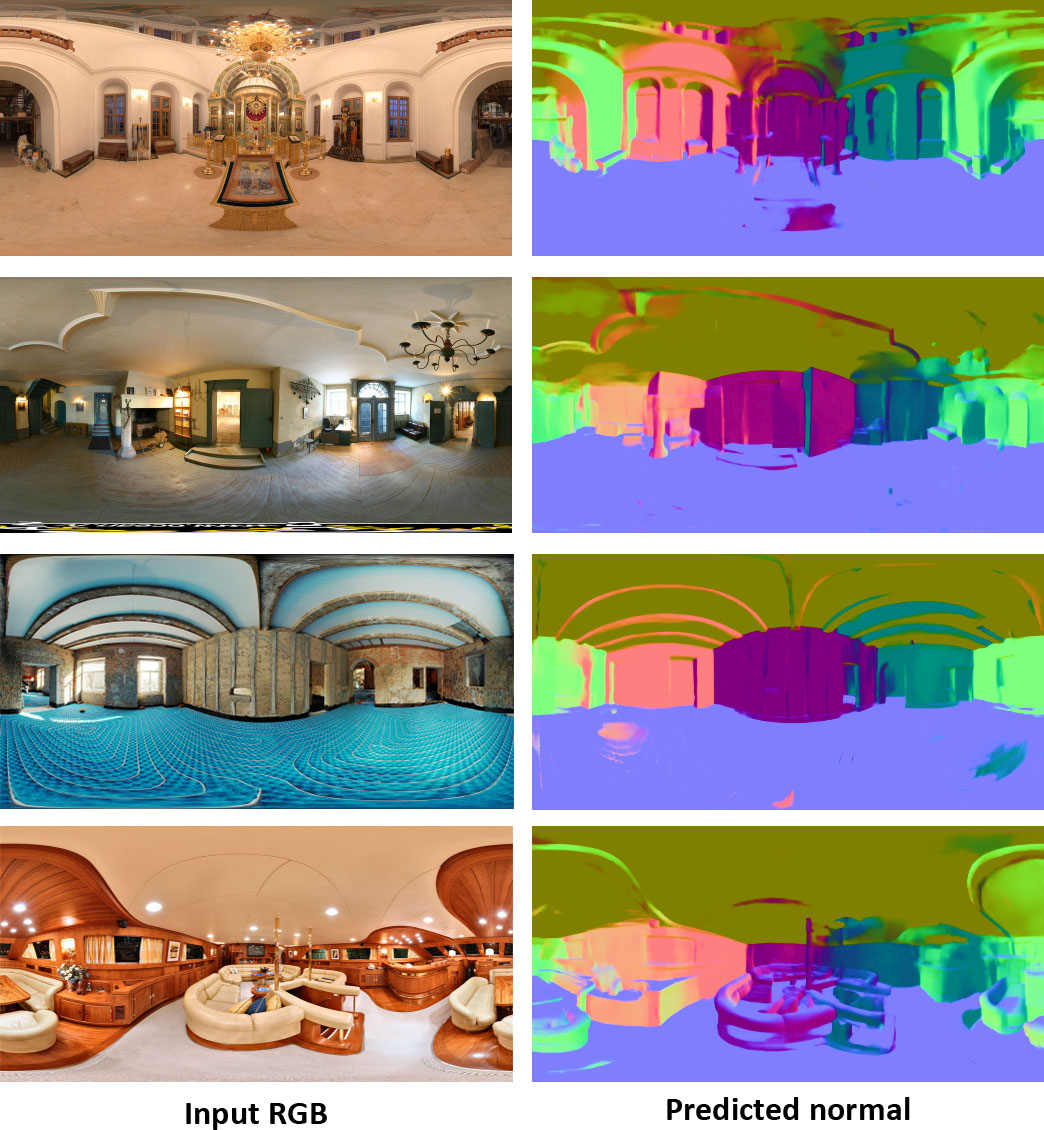}
    \caption{Qualitative results on samples of the realistic and unseen Sun360 \cite{xiao2012recognizing} that contains indoors scene panoramas.
    Our model infers valid surface estimates, even on these challenging scenes, even though trained on a mix of synthetic and real - but different distribution (i.e. saturation, lighting, content) - scenes.}
    \label{fig:sun360_qual}
\end{figure}
\begin{figure}[t]
    \centering
    \includegraphics[width = \linewidth]{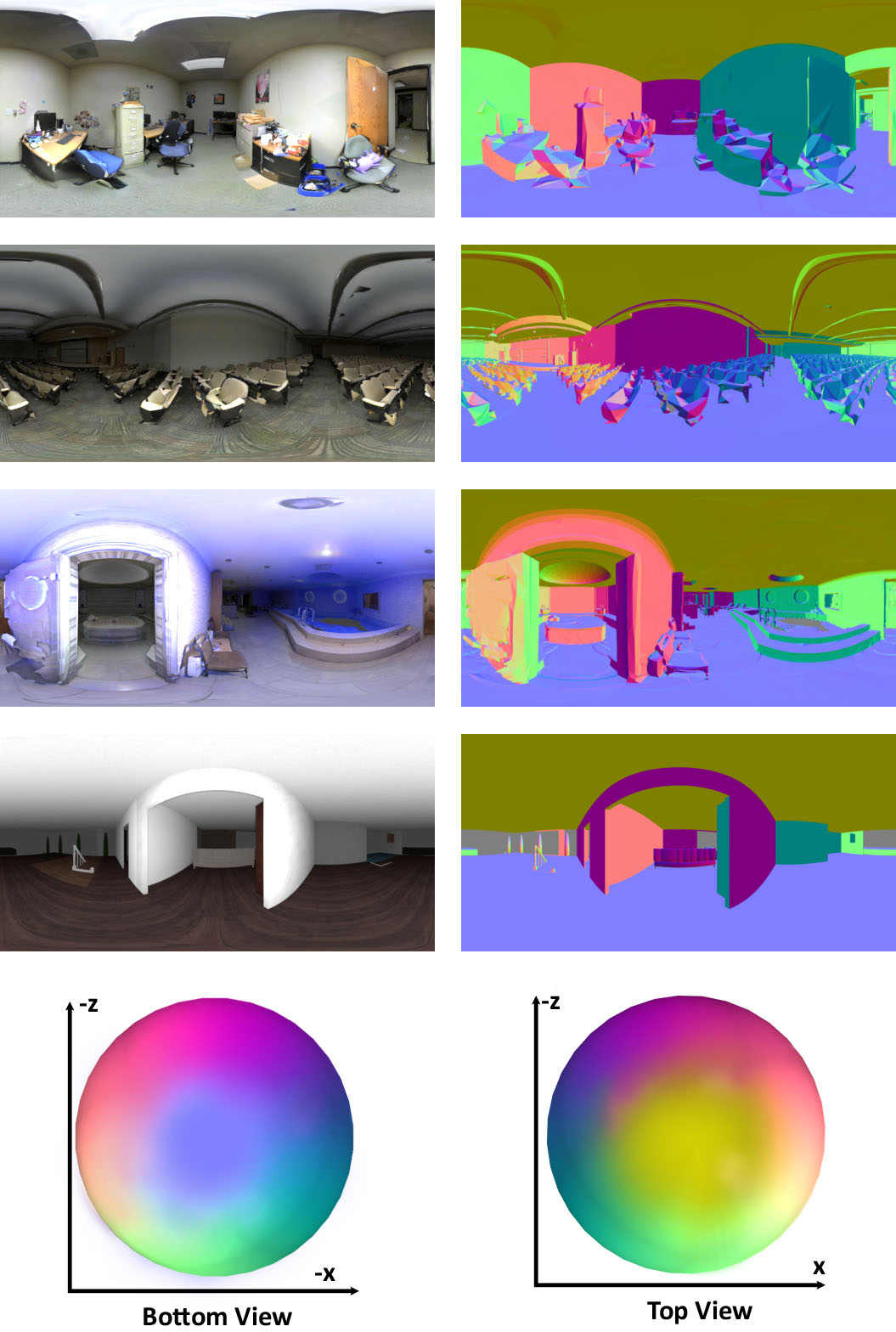}
    \caption{Samples of our generated dataset, with the color images next to their corresponding ground truth surface normal map. Bellow, the two color-spheres map pixel color to normal vector orientation. We consider a left-hand coordinate system, with the $z$ and $y$ axes representing the camera's look-up and up vectors respectively. Left: a bottom-view of the color-sphere, where the \textcolor{yUpColor}{center-color} represents the positive $y$-axis (\textcolor{yUpColor}{y+}). Right: a top-view of the color-sphere, where the \textcolor{yDownColor}{center-color} represents the negative $y$-axis (\textcolor{yDownColor}{y-}). The main coordinate axes \textcolor{xPos}{x+}, \textcolor{xNeg}{x-}, \textcolor{zPos}{z+}, \textcolor{zNeg}{z-}, have the same colors in both the color-spheres.} 
    \label{fig:dataset_showcase}
\end{figure}

Understanding 3D geometry from a single image is one of the most challenging and actively studied problems in computer vision. With the advent of efficient deep learning frameworks, many methods emerged that present state of the art results in tasks such as depth estimation \cite{laina2016deeper, godard2017unsupervised}, surface normal prediction \cite{fouhey2013data, wang2015designing} or a joint combination of both \cite{eigen2015predicting}. 3D visual perception can trace the path for a number of applications, like autonomous driving \cite{Zhan_2018_CVPR, Yin_2018_CVPR}, robot navigation \cite{zhu2017target}, 3D reconstruction \cite{avetisyan2019scan2cad} or even the fusion of two heterogeneous media, such as traditional 2D images with 3D objects for Augmented Reality (AR) applications \cite{nuernberger2016snaptoreality}.

Typical end-to-end deep learning pipelines usually require a large amount of ground truth annotated data. While this is partially addressed for datasets captured by traditional techniques following the typical pinhole camera projection model using depth sensors \cite{silberman2012indoor} or laser scanners \cite{saxena2008make3d}, the same cannot be said for \360 content \footnote{The terms \360, omnidirectional, spherical, equirectangular are equivalently used in this document.}, which is still considered a novel research domain with limited work done regarding 3D perception.

Nowadays, with good quality and efficient commercial based \360 spherical cameras and rigs, omnidirectional content is becoming increasingly popular and more easily produced. This expanded the usage of spherical sensors in a number of fields, such as Virtual Reality (VR) \cite{Lo:2017:VVD:3083187.3083219, Fan:2017:FPV:3083165.3083180}, indoor navigation \cite{adobeInv2018}, or even real-estate.

In this work we train a deep CNN on the task of single image \360 surface normal estimation. We address the lack of sufficient training data by generating a novel dataset of \360 indoors scenes with their corresponding ground truth surface annotations by rendering existing 3D datasets. The dataset is publicly available to enable further research in \360 visual perception \footnote{\url{vcl3d.github.io/HyperSphereSurfaceRegression}}. 

Inspired by the simplicity and numerical stability of quaternions when representing rotations, we train a deep CNN to predict \360 surface normal maps, by utilizing a novel loss function defined on the hyper-sphere using quaternions to express angular differences. Our experimental results (Table \ref{tab:ablation}) show  additional performance boost compared to models trained with losses commonly used in similar regression tasks. 

Additionally, we compare with other state of the art normal estimation methods trained on planar images by inferring their predictions on equirectangular as well as cubemap projections of our dataset's images. Finally, we present promising qualitative results of our network applied on completely unseen challenging samples of the Sun360 dataset \cite{xiao2012recognizing}, and further present the feasibility of our model for a \360 image-relighting application.
\section{Related Work}
\label{sec:related}

Since the goal of this work is to learn surface normal from a single \360 image, and to the best of our knowledge, similar work does not exist, we first present learning-based methods for the \360 domain, followed by similar work done on planar 2D datasets.

\subsection{Learning on \360 images}

The \360 field of view of spherical images benefits many applications, such as autonomous driving \cite{scaramuzza2008appearance}, robotics \cite{ran2017convolutional} or VR \cite{hu2017deep}. Typically, omnidirectional images are modeled as a sphere, and its pixel coordinates map to the longitudinal and latitudinal spherical coordinates. Despite their advantages, omnidirectional content suffers from distortion, especially near the sphere's poles, making it very difficult to process them with typical CNN architectures. Nowadays, the most usual ways to apply neural network pipelines on spherical input are either employing standard CNN architectures and run their predictions directly on the projected (typicaly equirectangular \cite{snyder1989album}) image, or projecting the image to the faces of a cube (cubemap) and then back-projecting them to equirectangular. However, there are a number of efforts that model the distortion of spherical images in the neural network's architectural processing pipeline.

To address spherical image distortion, many techiques utilized the gnomonic projection \cite{snyder1989album} to either model equirectangular distortion in the representation of the input data, or to guide convolution kernels' sampling pattern in order to learn distortion invariant features. In \cite{khasanova2017graph} a graph-learning approach for omnidirectional image classification is presented. The graph representing the image is constructed using the gnomonic projection and a method for designing convolutional kernels to have similar responses for the same patterns in the image regardless of the position and lens distortion is proposed. 
In a similar manner, a distortion aware convolution kernel sampling pattern is presented in \cite{tateno2018distortion}, which models the distortion in spherical images. The convolution kernels sample equirectangular images w.r.t. the gnomonic projection, and thus can be used with models trained on regular 2D images. SphereNet \cite{coors2018spherenet}, a framework for learning spherical image representations uses the same kernel sampling pattern, further boosting its computational performance by additionally sampling uniformly on the sphere using the method described in \cite{saff1997distributing}.

Other efforts, try to model equirectangular image distortion with more typical neural network architectures like \cite{su2017learning}, where the authors focus on learning to transfer trained 2D models to the spherical domain, by adjusting their network's kernel sizes w.r.t. to the latitudinal angle and enforcing consistency between the predictions of the 2D projected views and those in the \360 image. 

Additionally, there is limited work addressing 3D perception problems on the omnidirectional domain, such as \cite{DBLP:journals/corr/abs-1811-05304}, in which the authors follow the steps of \cite{zhou2017unsupervised}, to learn depth and camera motion from \360 videos, using two networks; one for inferring depth and the other for predicting the camera pose. They train their networks on cubemap projections of \360 video sequences rendered from the SunCG \cite{song2016ssc} dataset. Moreover, in \cite{zioulis2018omnidepth}, the authors use an end-to-end approach to learn \360 depth from equirectangular indoors scenes. They present a dataset generated via rendering existing 3D datasets and two neural network architectures, one more typical and the other constructed with rectangular filters and dilated convolutions \cite{yu2017dilated} to account for the distortion in the spherical domain.

Finally, in \cite{fernandez2019CFL} the authors focus on the task of learning a 3D room layout from a single \360 image, using the edges that are formed from wall-ceiling-floor intersections and their end-points, i.e. their corners, as their ground truth data. They achieve this by introducing equirectangular convolution kernels and a neural network trained on a subset of the Sun360 \cite{xiao2012recognizing} annotated with ground truth edge and corner data.

\begin{table*}[t]

\caption{Quantitative results of our model trained on our dataset's train-split and evaluated on our test-split with four different loss configurations. We present the mean, median and root mean square angular error across our dataset's test-set. We also provide an additional threshold of $5^{o}$ along with the most commonly used thresholds ($11.25^{o}$, $22.5^{o}$, $30^{o}$).  $\downarrow$ means lower is better, while $\uparrow$ means higher is better.}

\begin{center}
\begin{tabular}{l c | c c c | c c c c}
\hline
\hline
Network    & Loss       & Mean$\downarrow$      & Median$\downarrow$ & RMSE$\downarrow$ 
& $5^{o}\uparrow$ &$11.25^{o}\uparrow$    &  $22.5^{o}\uparrow$   & $30^{o}\uparrow$ \\ 

\hline

VGG16-UNet  & L2    & 7.72	& 7.23   & 8.39	& 73.55 & 79.88	& 87.72    & 90.43\\
VGG16-UNet  & Cosine    & 7.63	& 7.14   & 8.31	& 73.89 & 80.04	& 87.29    & 90.48\\
VGG16-UNet  & Quaternion & 7.24   & 6.72   & 7.98 & 75.8	& 80.59	& 87.3    & 90.37	\\
VGG16-UNet  & Quaternion + Smoothness   & \textbf{7.14}    & \textbf{6.66}    & \textbf{7.88} & \textbf{76.16}    & \textbf{80.82}   & \textbf{87.45}   & \textbf{90.47} \\

\hline
\hline

\end{tabular}
\end{center}

\label{tab:ablation}
\end{table*}

\subsection{Surface normal estimation from a single image}

The use of standard feedforward CNNs to predict a surface normal from a single RGB image has been employed by many recent works. Eigen and Fergus \cite{eigen2015predicting} propose a deep learning model for per-pixel regression using a sequence of three scales to generate features and refine predictions in a coarse to fine approach. Their network can be adapted to predict depth, surface normal or semantic segmentation by making small modifications to the architecture. 

In a more recent work \cite{zhang2017physically}, the authors introduce a synthetic dataset of indoors scenes, generated via physically-based rendering, with ground truth normal annotations, segmentation and object boundary masks. They pre-train a UNet \cite{ronneberger2015u} - VGG16 \cite{simonyan2014very} hybrid neural network model on their synthetic dataset and fine-tune it on NYUv2 \cite{silberman2012indoor}. A similar network architecture is adopted by \cite{bansal2016marr}, that presents an effort to retrieve 3D objects from 2D images. Their neural network is trained to predict surface normals that serve as input to another two-stream network that estimates the pose and the style of the depicted object in order to retrieve the object's 3D model from a large CAD library \cite{aubry2014seeing}.

One of the first approaches to propose a non-standard feedforward CNN architecture \cite{wang2015designing}, treats surface normal prediction as a classification problem instead of a regression one, based on \cite{ladicky2014discriminatively}. A three-model neural network architecture is presented which comprises a top-down, a bottom-up and a fusion network. The first learns a coarse global normal map and a room layout hypothesis incorporating vanishing point labels under a Manhattan World assumption. The second learns normals for a local patch of the input image and classifies the edges of the depicted scene as convex, concave and occlusion edges. Finally, the latter network fuses the predictions of the two input networks and outputs a final surface normal estimation of the input image.  

\subsection{Joint normal and depth estimation}

As depth and surface normals follow a strong geometric correlation \cite{shi2018planematch}, there are a number of methods that concentrate on learning surface normals and depth in a joint manner. Specifically in \cite{wang2016surge}, a four-branch neural network architecture that predicts dense depth and normals along with plane and edge probability maps is presented. The predictions are regularized by a dense conditional random field (DCRF) \cite{krahenbuhl2012efficient} that encourages the consistency of depth and normals within planar regions and enforces surface predictions to have unit length via the predicted edges and planes. 

Li \etal \cite{li2015depth}, use a pre-trained part of AlexNet \cite{krizhevsky2012imagenet} for depth estimation and VGG16 for surface prediction with non-trainable weights, which they feed with super-pixel patches of different sizes sampled from the input image. Their network makes as many predictions as the input patches, which are then concatenated and fed to two fully-connected layers with learnable parameters that produce the final depth (or surface normal) output. As a final refinement step they use a hierachical CRF that incorporates the relationship between the patches and the pixels of the image.   

In \cite{qi2018geonet}, Geonet is presented; a two-branch neural network trained to estimate depth and surface normal, using two new modules, the depth-to-normal and normal-to-depth networks, that both use pinhole camera geometry and the prediction of each branch to further refine the quality of their estimations.

An interesting method is presented in \cite{chen2017surface}, where the authors build on top of their previous work \cite{chen2016single}, and create a dataset by crowd-sourcing the annotation of images collected randomly from Flickr \footnote{\url{https://www.flickr.com/}}. They manage to train a neural network to estimate depth and surface normal using relative point-to-point depth and normal annotations evaluating their method on \cite{silberman2012indoor}.

Finally, the authors in \cite{le2018three} consider fusing two different sources of information other than depth with surface normal, namely optical flow and semantic segmentation, introducing a novel synthetic dataset of outdoor nature scenes, for general scene understanding. They show that joint features efficiency and the complementary refinement of one prediction from the other two, improves object boundaries and region consistency in predictions.

\section{Dataset Creation}

The data-driven nature of deep CNN architectures is partially addressed with datasets such as  \cite{silberman2012indoor} and \cite{saxena2008make3d}, for learning depth or surface normals given scenes captured by the pinhole camera projection model. However, it is difficult to obtain similar datasets of spherical images.

We overcome this limitation by following the steps of \cite{zioulis2018omnidepth}, and create a mixed dataset of spherical images of indoors scenes. Similarly, we used a path-tracing renderer \footnote{\url{https://www.cycles-renderer.org/}} and Blender \footnote{\url{https://www.blender.org/}} to render existing 3D datasets and annotate our rendered images with their corresponding ground truth surface normal maps that are produced as a result of the rendering process.

Specifically, we utilized the same 3D datasets, namely Matterport3D \cite{Matterport3D}, Stanford2D3D \cite{armeni20163d, armeni2017joint} and SunCG \cite{song2016ssc} to generate a dataset composed of a mixture of computer generated (CG) and realistic scenes of indoors spaces. The dataset consists of 24933 unique viewpoints, from which we split 7868 scenes for training, 1098 for validation and 2176 for benchmarking our trained models. We consider the remaining ones as invalid due to inaccuracies during rendering. We provide the dataset publicly to enable further research in \360 visual perception. We showcase a sample of our dataset in Fig. \ref{fig:dataset_showcase}. 

\section{\360 Surface Normals Estimation}
\label{sec:method}

Following most background work, we treat training a fully convolutional neural network (FCN) to learn surface normal from a single spherical image as a regression task. In most learning-based normal regression problems the approach is to minimize either the $L2$ norm \cite{li2015depth, qi2018geonet, bansal2016marr, dharmasiri2017joint} of the difference of the predicted normal map and the ground truth, or their normalized per-pixel dot-product \cite{eigen2015predicting, zhang2017physically} that implies their angular differences.

Quaternions can represent arbitrary rotations and surface orientation in a very simple and compact form.
To train our network, we consider normal vectors as pure quaternions and try to minimize their difference in terms of rotation, showing to further boost the performance of our model (Table \ref{tab:ablation}). 

We first formulate our novel quaternion loss function, followed by the description of the neural network architecture used for our experiments. 

\begin{figure}[h]
    \centering
    \includegraphics[width = \linewidth]{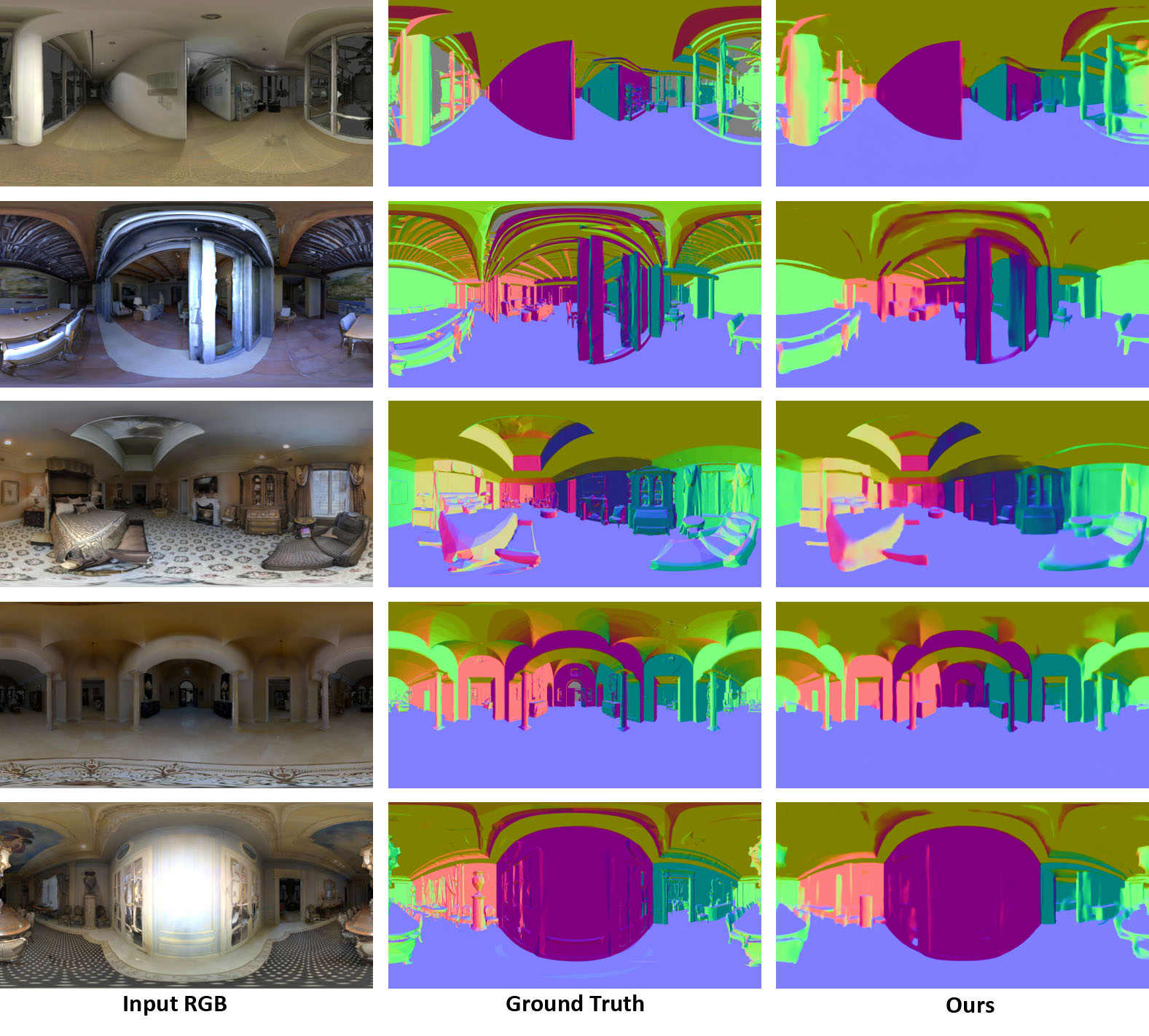}
    \caption{Qualitative results on samples of our test-split. From left to right: input equirectangular image, ground truth surface normal, our model's prediction.}
    \label{fig:qual_ours}
\end{figure}

\subsection{Angular loss on the hypersphere}

According to Euler's rotation theorem, a transformation of a fixed point 
$\textbf{p}(p_x, p_y, p_z) \in \mathbb{R}^3$ 
can be expressed as a rotation given by an angle $\theta$ around a fixed axis 
$\textbf{u}(x, y, z) = x\hat{\textbf{i}} + y\hat{\textbf{j}} + z\hat{\textbf{k}} \in \mathbb{R}^3$,
that runs through $\textbf{p}$. This kind of rotation is easily represented by a unit quaternion $\textbf{q}(w, x, y, z)$, where $w$ is the quaternion's real part, described by the following formula:

\begin{align}
    \textbf{q} &= e^{\frac{\theta}{2}(x\hat{\textbf{i}} + y\hat{\textbf{j}} + z\hat{\textbf{k}})} \implies \\ 
 \textbf{q} &= cos(\theta) + \textbf{u} sin(\theta) \label{eq:real_im_quat}
\end{align}
where $\|\textbf{q}\| = 1$, and $cos(\theta)$, $sin(\theta)$ are the quaternion's real and imaginary parts respectively.

Thereafter, we can represent two normal vectors 
$\hat{\textbf{n}_1}(n_{1_x}, n_{1_y}, n_{1_z})$ and 
$\hat{\textbf{n}_2}(n_{2_x}, n_{2_y}, n_{2_z})$  as the pure quaternions 
$\textbf{q}_1(0, n_{1_x}, n_{1_y}, n_{1_z})$ and 
$\textbf{q}_2(0, n_{2_x}, n_{2_y}, n_{2_z})$. 
Then, the angular difference between the two normal vectors can be expressed by their transition quaternion \cite{kuipers1999quaternions}, which represents a rotation from $\hat{\textbf{n}_1}$ to $\hat{\textbf{n}_2}$:

\begin{equation}
    \textbf{t} = \textbf{q}_1 \textbf{q}_2^{-1}
\end{equation}
Because $\textbf{q}_1$ and $\textbf{q}_2$ are unit quaternions $\textbf{q}^{-1} = \textbf{q}^*$,  where $\textbf{q}^*$ is the conjugate quaternion of $\textbf{q}$, and $\textbf{q}^{*} = -\textbf{q}$, due to being a pure quaternion, and:

\begin{equation}
    \textbf{q}_1 \textbf{q}_2^* = \textbf{q}_1 \cdot \textbf{q}_2 - \textbf{q}_1 \times \textbf{q}_2
\end{equation}

Therefore, because $\textbf{q}_1$ and $\textbf{q}_2$ are pure unit quaternions, their multiplication is reduced to a simple dot (real part) and cross product (imaginary part).
As a result, calculating the hypersphere angle represented by the transition quaternion can be straightforwardly implemented in most modern deep learning frameworks.  

The rotation angle of the transition quaternion $\textbf{t}$ and therefore the angular difference between the two normal vectors $\hat{\textbf{n}_1}$ and $\hat{\textbf{n}_2}$ is calculated by the inverse tangent between the real and the imaginary parts of the transition quaternion, which are reduced to their dot and cross product, due to being unit quaternions.

\begin{align}
     \tan{(\theta)} &= \frac{sin(\theta)}{\cos(\theta)} =  
     \frac{
        \|\textbf{q}_1 \times \textbf{q}_2 \|
     }
     {
        \textbf{q}_1 \cdot \textbf{q}_2
     } 
     \implies \\
    \theta &= atan(
        \frac{
            \| \textbf{q}_1 \times \textbf{q}_2 \|
        }
        {
            \textbf{q}_1 \cdot \textbf{q}_2
        }
    )
\end{align}
In the above computation, the only different operation against other typically used error functions, like the cosine similarity error, is the calculation of a cross product and the $atan()$ operator (we should note that in our implementation we use the $atan2()$ operator). However, these kind of operations are simple to implement and are supported by most deep learning frameworks. Additionally, this simplicity makes this loss function practical and with relatively low performance overhead.   

Due to imperfect scanning process, we do not consider invalid normals, during back-propagation by generating a mask $M(\textbf{p})$ at training time with its values being equal to zero for invalid pixels and one for the remaining ones.

Additionally, to further enhance our model's predictions on fine details and textureless regions, we add a weighted smoothness term (see \ref{subsec:smooth}) $E_{sm} = \|\nabla \Tilde{N}(\textbf{p})\|_2$ in the final error objective (for more information please refer to the supplementary material).

Finally, we minimize the following error:

\begin{equation}
\begin{split}
    E(\textbf{p}) = (1 - &\alpha)M(\textbf{p}) 
    \cdot 
    atan(
        \frac
        {
            \|\Tilde{N}(\textbf{p}) \times N(\textbf{p}) \|
        }
        {
            \Tilde{N}(\textbf{p}) \cdot N(\textbf{p})
        }
        ) \\
        &+ \alpha M(\textbf{p})\| \nabla\Tilde{N}(\textbf{p})\|
\end{split}
\end{equation}

\begin{table*}[h]
\begin{center}
\caption{Quantitative results against other monocular surface normal estimation models. Rows with \textit{equi} represent feeding the compared models with equirectangular images, and rows with $cube$ with cubemap projections following the method we describe in Sec. \ref{subsec:comparative}. }

\begin{tabular}{r l |c c|c c c c}

\cline{2-8}
\multicolumn{1}{r}{} & Network & Mean $\downarrow$ & Median$\downarrow$ & $5^{\circ}\uparrow$ &$11.25^{\circ}\uparrow$ & $22.5^{\circ}\uparrow$ & $30^{\circ}\uparrow$ \\ 
\cline{2-8}
\multicolumn{1}{r}{} & VGG16-UNet &$ \textbf{7.14}$	& $\textbf{6.66}$	&$ \textbf{76.16}$	& $ \textbf{80.82}$	& $ \textbf{87.45}$ & $\textbf{90.47}$	\\
\cline{2-8}
\multirow{2}{*}{\rotatebox{90}{equi}} & Zhang \etal \cite{zhang2017physically} & $41.85$	& $27.76$	&$11.4$	&$31.5$	&$45.2$	& $51.8$	\\
\multicolumn{1}{r}{}& Chen \etal \cite{chen2017surface}  & $51.37$	& $38.29$	&$2.7$	&$11.8$	&$31.0$	& $40.8$	\\    
\cline{2-8}
\multirow{2}{*}{\rotatebox{90}{cube}}& Zhang \etal \cite{zhang2017physically} & $26.12$	& $20.83$	&$9.1$	&$26.9$	&$53.3$	& $66.0$	\\
\multicolumn{1}{r}{}& Chen \etal \cite{chen2017surface}  & $27.10$	& $19.42$	&$6.2$	&$25.9$	&$56.0$	& $68.9$	\\
\cline{2-8}     
\label{tab:quantitative_comp}
\end{tabular}
\end{center}

\end{table*}

\subsection{Loss function comparison}

\begin{figure}[h!]
    \centering
    \includegraphics[width = \columnwidth]{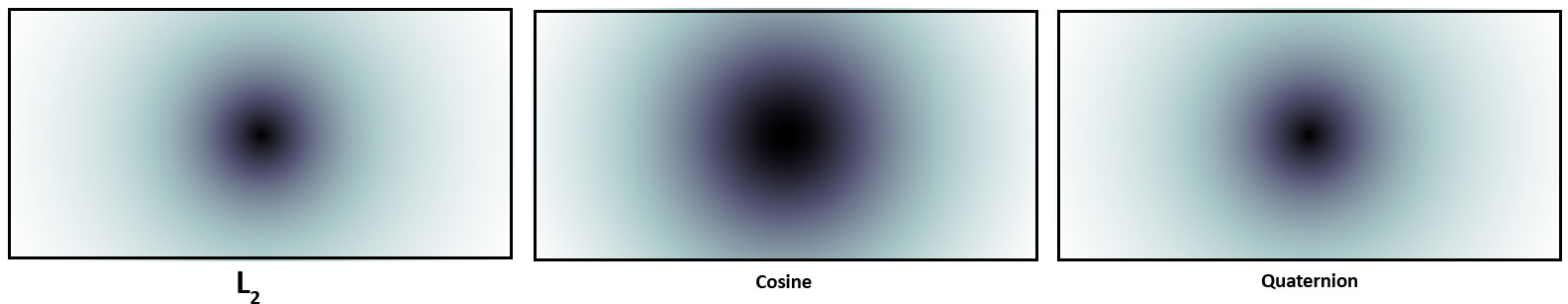}
    \caption{2D loss error landscape of the commonly used $L_2$ and cosine loss functions compared with our presented quaternion error function. From left to right: $L_2$, cosine and quaternion. (Darker color means higher error).}
    \label{fig:losses2d}
\end{figure}

\begin{figure}[h!]
    \centering
    \includegraphics[width = \columnwidth]{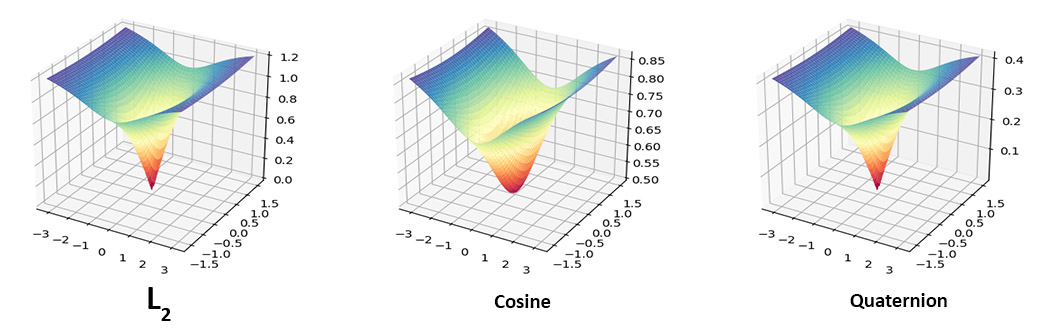}
    \caption{3D loss error landscape of the commonly used $L_2$ and cosine loss functions compared with our presented quaternion error function. From left to right: $L_2$, cosine and quaternion.}
    \label{fig:losses3d}
\end{figure}

In order to understand the efficiency of our novel quaternion loss function we provide 2D and 3D plots of its error landscape comparing them with the ones of the commonly used $L_2$ and \textit{cosine} loss functions. The 2D plots are presented in Fig \ref{fig:losses2d} while the 3D ones in Fig. \ref{fig:losses3d}.

In order to calculate the different error functions' landscapes we consider a reference vector $\Vec{n}$ and calculate the error between all the vectors generated in a $512 \times 256$ grid and the reference one, using each compared error function. 

The 2D plots in Fig \ref{fig:losses2d} do not provide much information about the nature of the compared error functions. However, it is obvious that the error landscape of the quaternion loss function is more convex than the cosine one. Thus, we can justify the better performance of our network using the quaternion loss function. In addition, we should note that despite the error landscape of the $L_2$ loss function appears to have similar convexity with the quaternion one, being just a difference between two values, it does not incorporate the three dimensional nature that is needed to solve 3D learning-based problems.


\subsection{Smoothness term}
\label{subsec:smooth}
\begin{figure*}
    \centering
    \includegraphics[width = \linewidth]{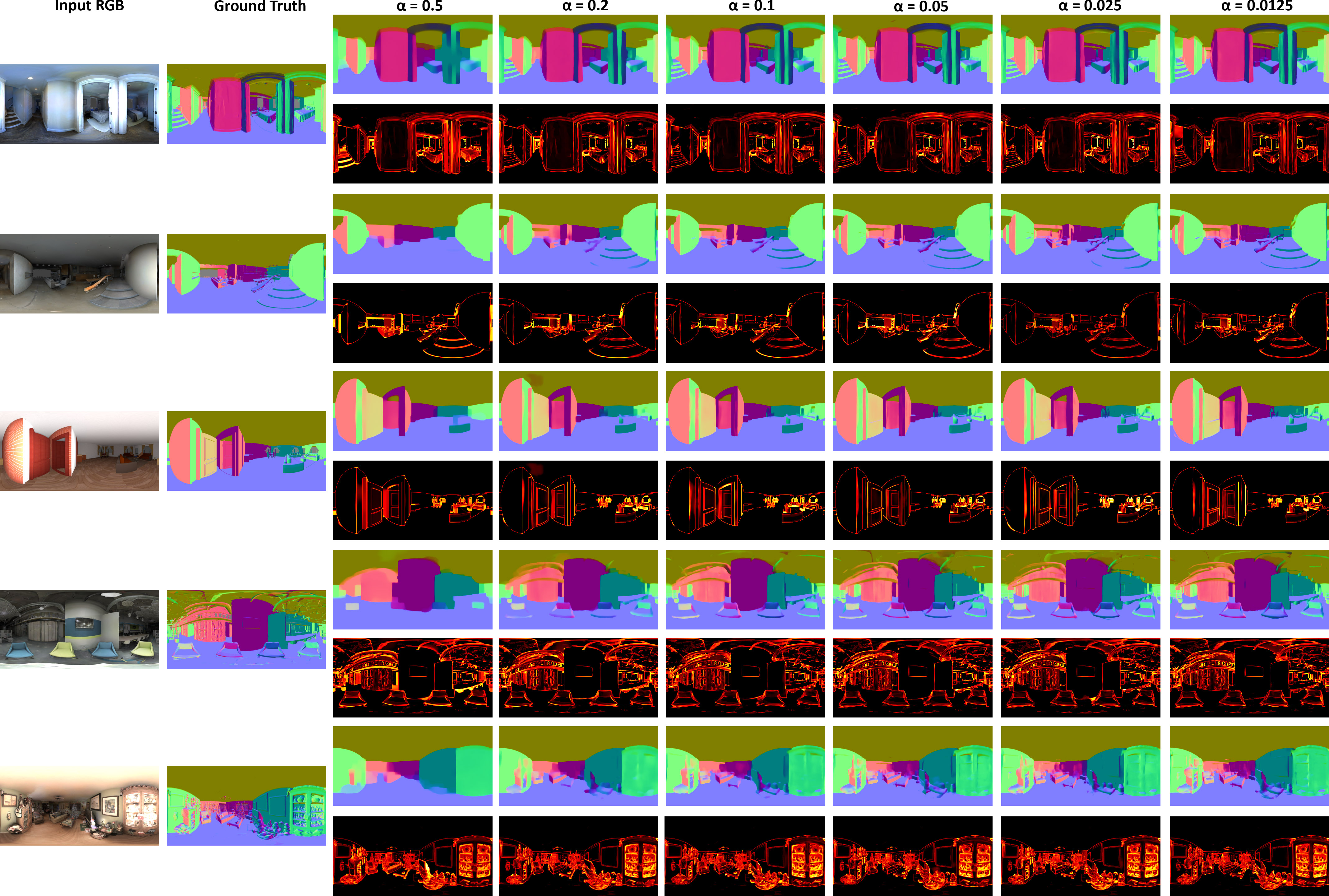}
    \caption{Qualitative results on a sample of our test-split for different loss weighting factors $\alpha$. From left to right: The input color image, its corresponding ground truth surface normal, predictions for $\alpha = 0.5, 0.2, 0.1, 0.05, 0.025$ and $0.0125$, with their respective error maps underneath each prediction.}
    \label{fig:sup_qual_alpha}
\end{figure*}

Following the works of \cite{heise2013pm, godard2017unsupervised, zioulis2018omnidepth}, we use a gradient smoothing term, i.e. an $L_{1}$ error on the prediction's gradients, in the final loss function. As surface normal discontinuities often occur at image gradients, this encourages our network's predictions to be locally smooth, penalizing gradients that may wrongly occur from the texture of the input image. We consider the weighting term $\alpha$ as a hyper-parameter, and experimentally choose $\alpha = 0.025$ in our final implementation. In Table \ref{tab_sup:alpha_ablation} we provide our network's results on our test-set for range of different $\alpha$ values. Additionally, in Fig. \ref{fig:sup_qual_alpha} we present qualitative results on a sample of our dataset's test-split.

From the results in Fig. \ref{fig:sup_qual_alpha} we can conclude that high $\alpha$ values result in over-smoothed predictions with very little level of detail. Additionally, when $\alpha$ is set to a very low value it has very little impact on the training of the model.

\begin{table*}[t]

\caption{Quantitative results of our model trained on our dataset's train-split and evaluated on our test-split for different values of the loss weighting factor $\alpha$. We also provide an additional threshold of $5^{o}$ along with the most commonly used thresholds ($11.25^{o}$, $22.5^{o}$, $30^{o}$).  $\downarrow$ means lower is better, while $\uparrow$ means higher is better.}

\begin{center}
\begin{tabular}{l | c c c | c c c c}
\hline
\hline
$\alpha$       & Mean$\downarrow$      & Median$\downarrow$ & RMSE$\downarrow$ 
& $5^{o}\uparrow$ &$11.25^{o}\uparrow$    &  $22.5^{o}\uparrow$   & $30^{o}\uparrow$ \\ 

\hline

$\alpha = 0.5$    & 8.61	     & 8.12         & 9.41          & 75.4          & 79.44         & 85.4          & 88.19\\
$\alpha = 0.2$    & 7.38	     & 6.84         & 8.13          & 76.21         & 80.62	        & 87.02         & 89.99\\
$\alpha = 0.1$    & 7.2          & 6.7          & 7.93          & \textbf{76.18}& \textbf{80.84}& 87.35         & 90.36	\\
$\alpha = 0.05$   & 7.18         & \textbf{6.66}& 7.91          & 76.12	        & 80.78	        & 87.38         & 90.41	\\
$\alpha = 0.025$  & \textbf{7.14}& \textbf{6.66}& \textbf{7.88} & 76.16         & 80.82         & \textbf{87.45}& \textbf{90.47} \\
$\alpha = 0.0125$ & 7.17         & 6.65         & 7.89          & 76.04	        & 80.78	        & 87.4          & 90.36	\\

\hline
\hline

\end{tabular}
\end{center}

\label{tab_sup:alpha_ablation}
\end{table*}

\subsection{CNN architecture}
\label{subsec:architecture}

Adopting the work of \cite{zhang2017physically, bansal2016marr}, we utilize a fully convolutional (FCN) \cite{long2015fully} encoder-decoder network with skip-connections that regresses towards the ground truth surface normals. The network architecture is based on UNet \cite{ronneberger2015u} combined with a VGG16 \cite{simonyan2014very} encoder. Despite, training other models used in the literature, their performance was inferior to the selected architecture.  

Typically a UNet architecture consists of an encoder that captures the input image's context, and a symmetrical decoder that enables precise localization. In our implementation, the front-end encoder remains the same as $conv1$-$conv5$ in VGG16, and the decoder is composed of symmetrical blocks of convolutions and bi-linear up-sampling layers. In order to localize the decoder's upsampled features, we concatenate them with their symmetrical high-resolution features from the encoder via skip-connections. This technique is shown to prevent gradient degradation \cite{he2016deep}, and proved to be an important element in the network's design. Our model outputs high resolution results and keeps fine object details that might otherwise disappear between pooling and up-sampling layers. 

Further, we use ReLU \cite{nair2010rectified} as the activation function and batch normalization \cite{ioffe2015batch} after each convolutional layer. Finally, the output of the network is fed to a convolution with a $3\times3$ kernel size to produce the final 3-channel prediction, which we explicitly normalize along the channel dimension.

\begin{figure*}[!th]
\centering
    \begin{tabular}{c}
    \rotatebox{90}{\scriptsize{Input rgb}}
    \includegraphics[width = 0.9\textwidth]{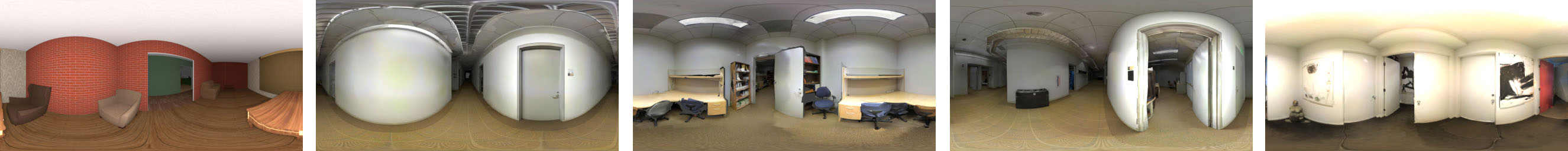} 
    
    \\
    \rotatebox{90}{\scriptsize{Ground Truth}}
    \includegraphics[width = 0.9\textwidth]{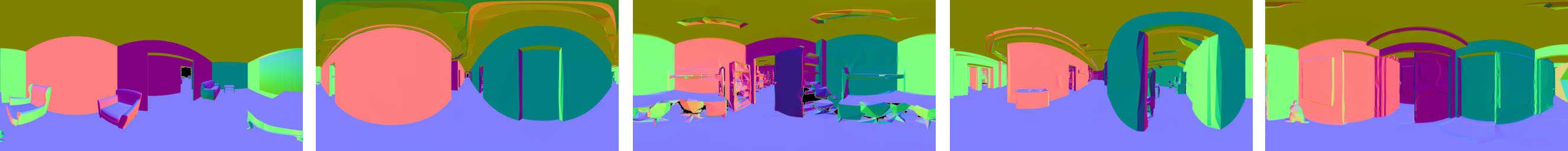} 
    
    \\
    \rotatebox{90}{\scriptsize{Chen \etal \cite{chen2017surface}}}
    \includegraphics[width = 0.9\textwidth]{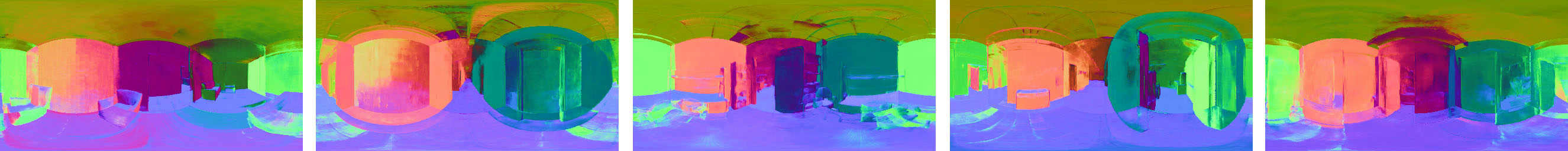} 
    
    \\
    \rotatebox{90}{\scriptsize{Zhang \etal \cite{zhang2017physically}}}
    \includegraphics[width = 0.9\textwidth]{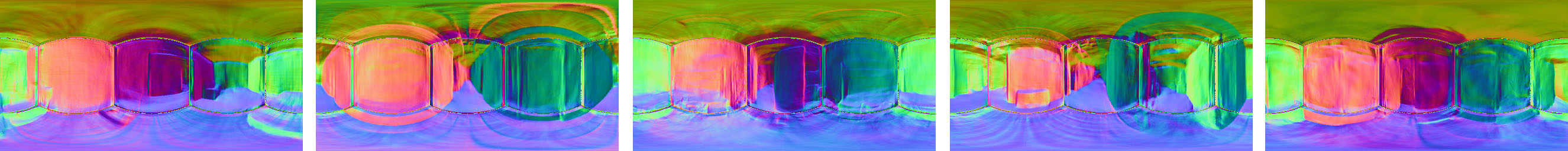} 
    
    \\
    \rotatebox{90}{\scriptsize{Ours}}
    \includegraphics[width = 0.9\textwidth]{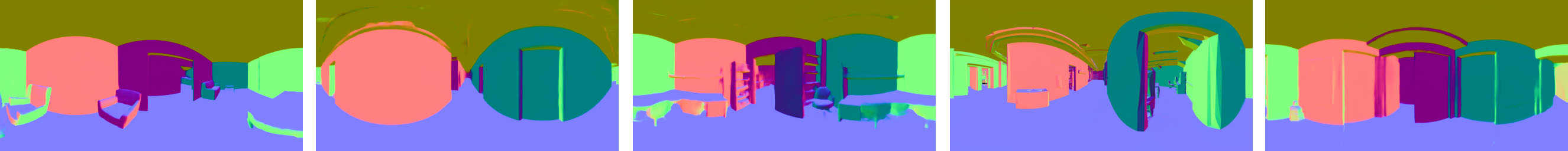} 
    
    \end{tabular}    
    \caption{Qualitative comparisons against \cite{chen2017surface} and \cite{zhang2017physically}. We present results for a sample of our test-split by applying the compared models to cubemaps of our dataset following the method we describe in Sec. \ref{subsec:comparative}}
    \label{fig:qual_comp}
\end{figure*}

\section{Experimental Results}
\label{sec:results}
This section provides an experimental evaluation of our method. To assess the efficiency of our quaternion loss function, we first train our model using the $L_2$ norm of the difference of the predicted and the ground truth surface normal, and additionaly, with their normalized per-pixel dot product, i.e. their cosine similarity. We then compare their performance on our dataset's test split. 

We then evaluate its performance compared to other methods applied on cubemap projections of our dataset as well as the original equirectangular images. 

Additionally, we show the efficacy of our model's generalization ability, by applying it on a subset of the Sun360 dataset containing unseen indoors scenes. Our trained model produces very promising qualitative results, even on in-the-wild data coming from considerably different distributions from our dataset's train-split. To further evaluate its effectiveness, we experiment with an image relighting application \cite{Ramamoorthi:2001:ERI:383259.383317}. We compare relit images using our model's predictions to relight them, and present qualitative results on samples of our dataset and a subset of Sun360.

\subsection{Training Details}
\label{subsec:train_details}

All of our networks were implemented and trained using pyTorch \cite{paszke2017automatic} framework. Experiments were performed on a PC equipped with an NVIDIA TITAN X GPU, CUDA \cite{nickolls2008scalable} v9.0 and and CuDNN \cite{chetlur2014cudnn} v7.1.3. We used a random seed of 1337 for all of our experiments, for achieving similar training sessions and reproducibility. We initialize our network's encoder parameters with weights pre-trained on ImageNet \cite{imagenet_cvpr09}, and the remaining convolution layers with Xavier weight initialization \cite{glorot2010understanding}. We use ADAM \cite{kingma2014adam} as the optimizer with its default parameters $[\beta_1, \beta_2, \epsilon] = [0.9, 0.999, 10^{-8}]$ and a learning rate of $0.0002$, and we train all of our models for $50$ epochs. We feed every network with equirectangular images at a $512 \times 256$ resolution, with the models' predictions being of equal size. Finally, we use a loss weighting factor $\alpha = 0.025$ between the prediction and the smoothness term.

\begin{figure}
    \centering
    \includegraphics[scale = 0.15]{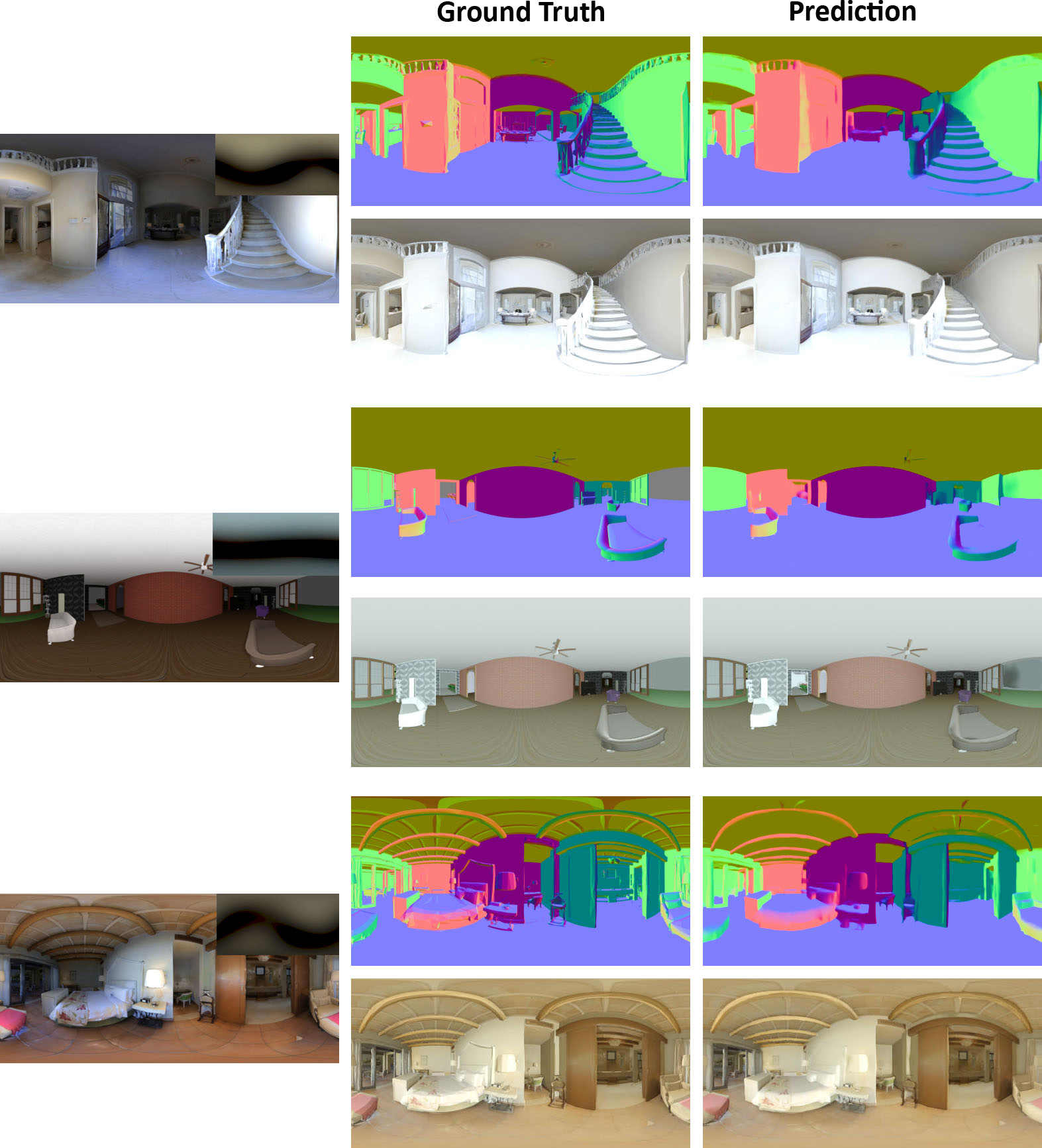}
    \caption{Qualitative comparison of images relit using ground truth normals and our model's prediction. For each row, the second column presents the ground truth normal map and the relit image, while the second, our model's prediction and the relit result. In addition, the irradiance map that used to relight both of the images is provided as an inset in the Input rgb image.}
    \label{fig:relit}
\end{figure}
\subsection{Model Performance}
\label{subsec:modelresults}

To evaluate our results, we use well-established error metrics that are described in the literature, initially introduced in \cite{fouhey2013data}. We measure the mean, median and root-mean-square (RMSE) angular error between the predicted and ground truth normal maps across our dataset's test split. Furthermore, we present precision coverage errors for three commonly used thresholds, namely $11.25^o, 22.5^o$ and $30^o$ and additionally $5^{o}$.

Table \ref{tab:ablation} presents the results of our model evaluated on our test-set when trained under four different loss function configurations, while in Fig. \ref{fig:qual_ours} we provide qualitative results of our best performing model. First, we can observe that the models trained with a more intuitive loss function that incorporates geometric understanding, like the cosine similarity or the quaternion loss, have improved performance over the one trained using a generic loss function like the $L_2$ norm. Additionally, the model trained with our proposed quaternion error outperforms all the others, with the results getting further improved when we add a smoothness term in the loss function.

\begin{figure}[t]
    \centering
    \includegraphics[width = \linewidth]{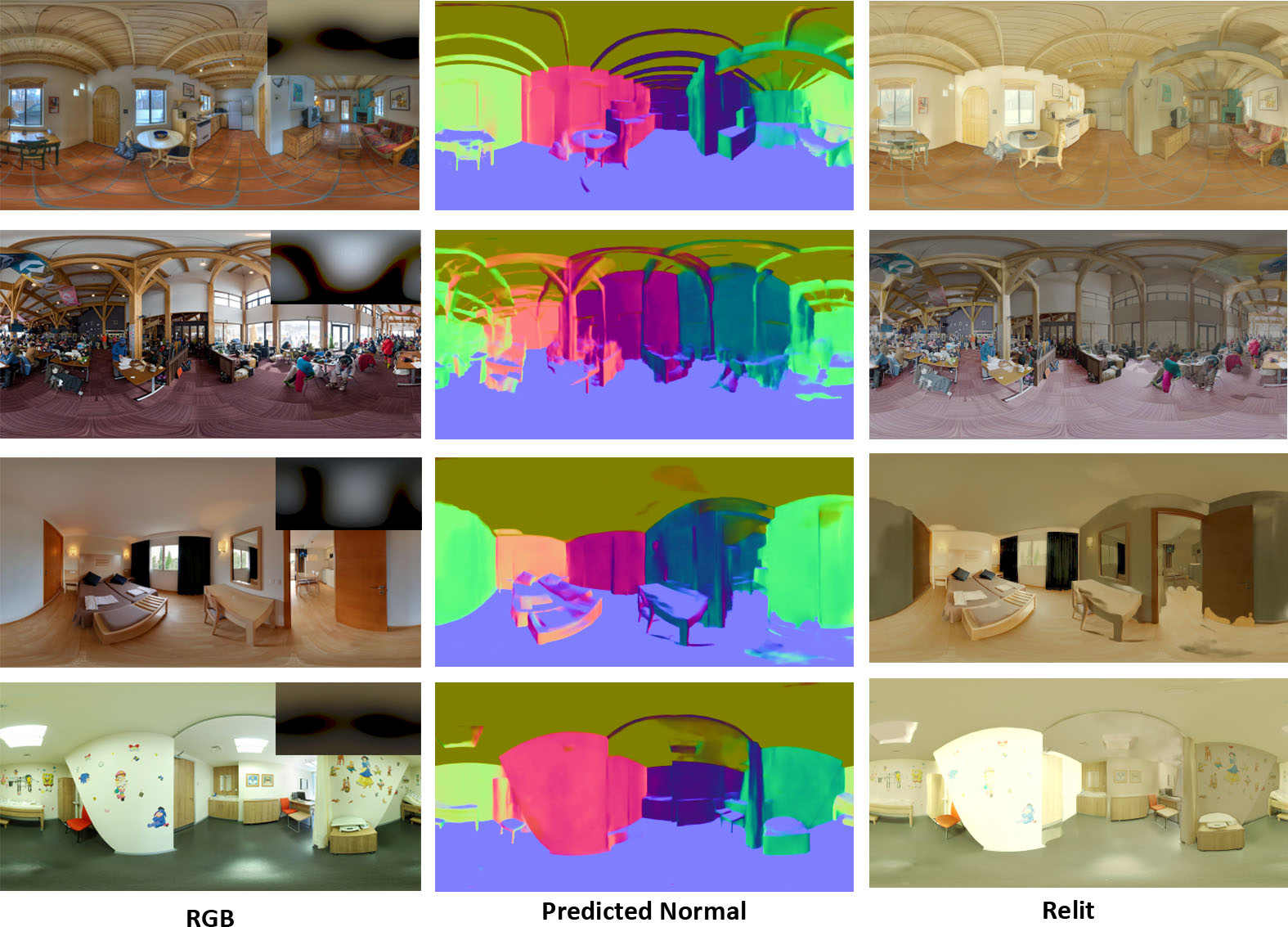}
    \caption{Samples from the Sun360 dataset, relit using our model's predictions. For every sample we provide the input rgb, our model's surface prediction and the relit image. The irradiance map used to relight the images is provided as an inset in the input rgb image.}
    \label{fig:sun360_relit}
\end{figure}
\begin{table*}[h!]

\caption{Quantitative results of our model trained on our dataset's train-split and evaluated on our test-split compared to the two neural network architectures for omnidirectional monocular depth estimation presented in \cite{zioulis2018omnidepth} and the method of \cite{zhang2017physically} re-trained on our dataset's train-set. We present the mean, median and root mean square angular error across our dataset's test-set. We also provide an additional threshold of $5^{o}$ along with the most commonly used thresholds ($11.25^{o}$, $22.5^{o}$, $30^{o}$).  $\downarrow$ means lower is better, while $\uparrow$ means higher is better.}

\begin{center}
\begin{tabular}{l | c c c | c c c c}
\hline
\hline
Network      & Mean$\downarrow$      & Median$\downarrow$ & RMSE$\downarrow$ 
& $5^{o}\uparrow$ &$11.25^{o}\uparrow$    &  $22.5^{o}\uparrow$   & $30^{o}\uparrow$ \\ 

\hline

Ours                                    & \textbf{7.14}	& \textbf{6.66}   & \textbf{7.88}	& \textbf{76.16} & \textbf{80.82}	& \textbf{87.45} & \textbf{90.47}\\
UResNet \cite{zioulis2018omnidepth}     & 29.86	& 30.1   & 30.21& 9.91  & 25.18	& 48.84 & 60.42\\
RecNet \cite{zioulis2018omnidepth}      & 31.64 & 31.95  & 32.1 & 8.24  & 21.26	& 44.19 & 56.59\\
Zhang \etal \cite{zhang2017physically}  & 11.03& 10.61& 11.72& 62.9& 73.88& 82.74& 86.56 \\ 

\hline
\hline

\end{tabular}
\end{center}

\label{tab:sup_omnidepth}
\end{table*}

\begin{figure*}[t!]
    \centering
    \includegraphics[width = \linewidth]{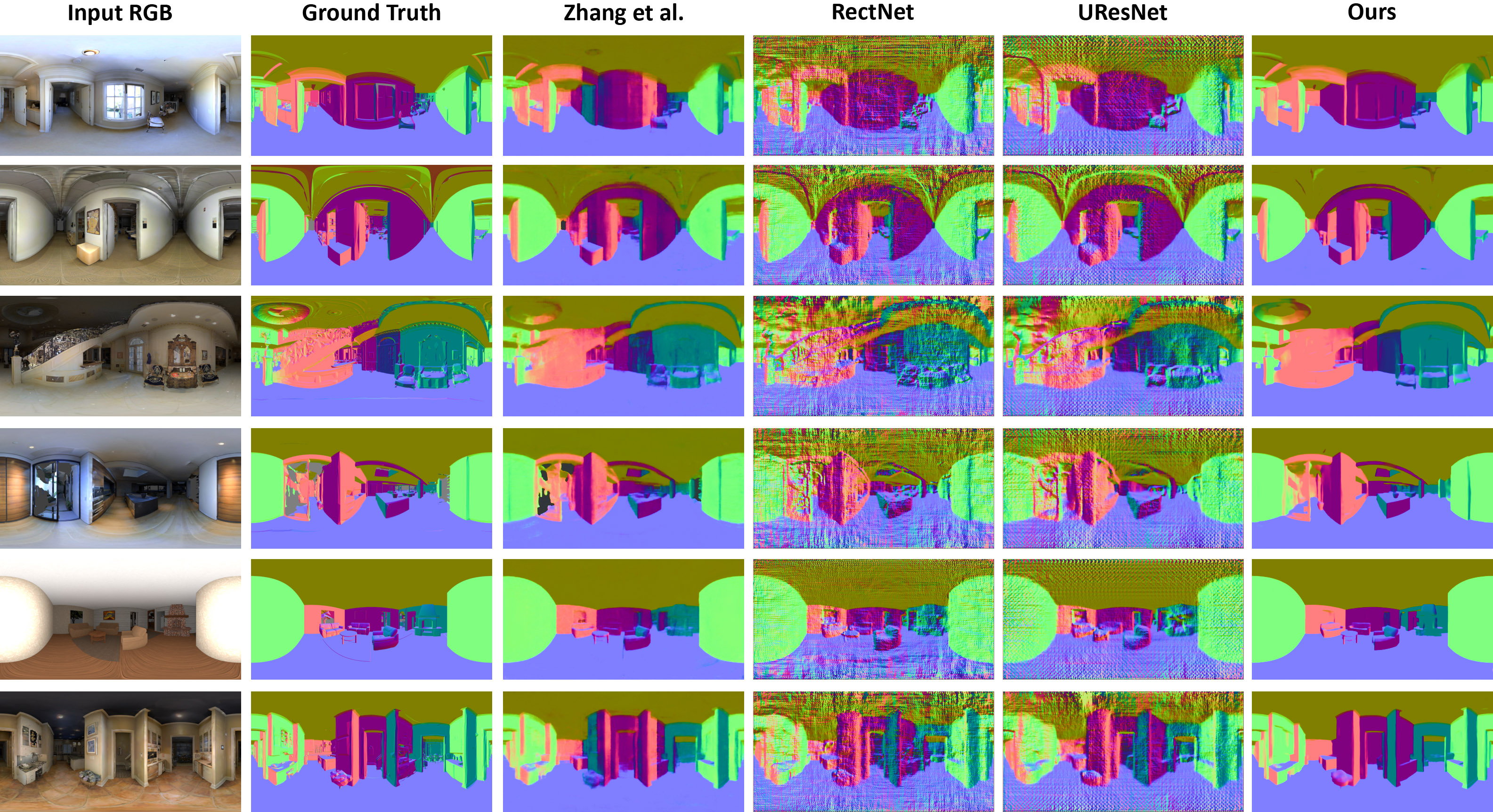}
    \caption{Qualitative comparison on a sample of our test-set. From left to right: Input RGB, ground truth surface normal, Zhang \etal \cite{zhang2017physically} re-trained on our test-set, RectNet \cite{zioulis2018omnidepth}, UResNet \cite{zioulis2018omnidepth}, our model.}
    \label{fig:sup_qual_omnidepth}
\end{figure*}

\subsection{Comparison against other methods}
\label{subsec:comparative}

To the best of our knowledge, there is no other similar work on monocular \360 surface normal estimation. In an effort to show the importance of training directly on the omnidirectional domain, we provide comparisons of our model with learning-based methods trained on traditional perspective images. Specifically, we employ \cite{zhang2017physically}, which utilizes a similar neural network architecture, and \cite{chen2017surface} which is trained on a dataset with relative depth and surface normal annotations.

To accomplish this in a fair manner, we follow two schemes. First, we run the predictions of the compared models directly on the equirectangular images of our dataset's test-split, to evaluate how well 2D learned features cope with the distortion on the spherical domain. Moreover, we feed them cubemap projections of spherical images, and exploiting the known rotations between the cube's faces, we rotate the predicted normal vectors accordingly when back-projecting them to equirectangular. We should note that because our dataset is composed of indoors scenes, the top and bottom faces of the cubemap projections depict only portions of ceiling and floor content respectively. These mostly contain equally textured areas, not sufficient for detecting features in an image. Thus, we do not consider these areas when measuring each model's performance by masking them in the final error computation. 

In addition, our dataset contains floor-aligned camera poses, which is in contradiction with the datasets used to train the compared methods. These datasets contain scenes captured by arbitrary camera poses not necessarily aligned to the floor. Thus, models trained on them would possibly make rightful predictions but unaligned to our dataset's global orientation. To account for that, we perform singular value decomposition (SVD) between the prediction and the ground truth, and apply the resulting rotation to the prediction, before we calculate their error.

Results of both our evaluation methods are presented in Table \ref{tab:quantitative_comp}, and qualitative samples in Fig. \ref{fig:qual_comp}. When we run the compared models on cubemaps instead of directly on spherical images, both of the networks' performance is superior. This is expected, as these models are trained on 2D datasets and cannot produce effective features from the characteristics of the distorted equirectangular images. However, we can clearly see discontinuities and inconsistency between each cubemap face. We associate this to the fact that a $90^o$ FOV camera cannot capture global context information required for the models to make consistent predictions.

Finally, we compare with the omnidirectional depth estimation models presented in \cite{zioulis2018omnidepth}, by running their predictions on our test-set and then converting the resulted $360^{o}$ depth maps to surface maps. Additionally, we re-train the model of \cite{zhang2017physically} on our train set. Qualitative results are presented in Table \ref{tab:sup_omnidepth}, with additional qualitative results presented in Fig. \ref{fig:sup_qual_omnidepth}.

\subsection{Surface normal estimation and \360 scene relighting}

To further evaluate the performance of our model, we additionally experiment with spherical image re-lighting. We examine \cite{Ramamoorthi:2001:ERI:383259.383317}, in which the authors focus on rendering diffuse objects lit from a given environment map.
They show that the scene's irradiance, being a function of the the scene surface normal only, can be approximated in terms of a quadratic polynomial incorporated in the cartesian coordinates of the normal vector, by only 9 spherical harmonic coefficients with an error of only $1\%$. Specifically, the final relit image is composed of a sum of spherical harmonic basis functions, scaled by the lighting coefficients of the given environment map.

To extract natural spherical harmonic coefficients, we use a dataset of HDR indoors environment maps introduced in \cite{lalonde}. We utilize 9 lighting coefficients for relighting our images, which are later used for estimating an analytic approximation. Finally, the irradiance scaled by each pixel's intensity produces the output relit image.

We provide qualitative results of images sampled from our test-set in Fig \ref{fig:relit}, and additionally, in Fig \ref{fig:sun360_relit}, we present samples from Sun360. The first, are relit using both the ground truth normals and predictions of our model, while the second only with our model's output. Again our network shows promising results, as the differences between the two reilit images are almost imperceptible, and manifest mostly in highly detailed regions of the image.
\section{Conclusion \& Discussion}

In conclusion, we address the task of monocular \360 surface estimation as a learning problem. To overcome the lack of sufficient training data, we resolve to leveraging 3D rendering to generate spherical images of synthetic (CG) as well as realistic 3D datasets, along with their respective ground truth normal maps and make this dataset publicly available online. In addition, we train a deep CNN to estimate spherical surface normal given a single equirectangular image as input, by employing a simple to implement novel loss function. Our results show better network performance when it is trained with our proposed error function. Furthermore, they demonstrate that when 3D perception is assimilated in the learning objective, neural networks that tackle 3D geometry problems achieve better results. Additionally, we qualitatively present the generalization ability of our trained model via running its predictions on in-the-wild data and using them for an image re-lighting application. 

3D perception on spherical media is still considerably unexplored despite their wide utilization. Synthesizing data to circumvent the lack of spherical datasets can be a solution for training neural network models. However, these data will be product of rendering CG or large-scale scanned 3D models, that contain inaccuracies and invalid information. Additionally, it is very difficult to cover a large amount of real-life indoors or outdoors scenes. Accounting for the disadvantages of synthetic data, in the future, we would like to experiment with 3D perception on arbitrary \360 video sequences, employing self-supervised deep neural network models, and additionally model the spherical distortion in the neural network's architecture.

\section*{Acknowledgements}
This work was supported and received funding from the European Union Horizon 2020 Framework Programme 
project Hyper360, under Grant Agreement no. 761934. We are also grateful and acknowledge the support of NVIDIA for a hardware donation



{\small 
\bibliographystyle{ieee}
\bibliography{egbib}
}

\end{document}